\documentstyle[twoside]{siam}
\input{psfig}

\newcommand{\cond}{{\rm cond}}

\renewcommand{\span}{{\rm span}}
\newcommand{\diag}{\, {\rm diag}}

\newdimen\pIR
\newcommand{\R}{{\cal R}}

\renewcommand{\theequation}{\arabic{section}.\arabic{equation}}

\begin{document}

\title{Principal Manifolds and Nonlinear Dimension\\[3pt]
Reduction via Local Tangent Space Alignment}
\author{
Zhenyue Zhang\thanks{
Department of Mathematics, Zhejiang University, Yuquan Campus,
Hangzhou, 310027, P. R. China. {\tt zyzhang@math.zju.edu.cn}.
The work of this
author 
was done while visiting Penn State  University and was supported in part
by the Special Funds for Major State Basic Research Projects (project
G19990328), Foundation for University Key Teacher by the
Ministry of Education, China, and
NSF grants CCR-9901986. }
\and
Hongyuan Zha\thanks{Department of
Computer Science and Engineering, The  Pennsylvania  State University,
University Park, PA 16802, {\tt zha@cse.psu.edu}. The work of this
author
was supported in part by
NSF grants CCR-9901986.}
}

\maketitle

\markboth{ZHENYUE ZHANG AND HONGYUAN ZHA}{
Principal Manifolds via Local Tangent Space Alignment
}
\pagestyle{myheadings}

\begin{abstract}
Nonlinear manifold learning from unorganized data points is a very
 challenging unsupervised learning 
and data visualization problem with a great variety of
 applications. In this paper we present a new algorithm for manifold learning
 and nonlinear dimension reduction. Based on a set of unorganized data points
 sampled with noise from the manifold, we represent the local geometry of the
 manifold using tangent spaces learned by fitting an affine subspace in a
 neighborhood of each data point. Those tangent spaces are aligned to give
 the internal global coordinates of the data points with respect to the
 underlying manifold by way of a partial eigendecomposition of the
 neighborhood connection matrix. We present a careful error analysis of our
 algorithm and show that the reconstruction errors are of second-order
 accuracy. We illustrate our algorithm using curves and surfaces both in
 2D/3D and higher dimensional Euclidean spaces, and 64-by-64
 pixel face images with various pose and lighting conditions. We also address
 several theoretical and algorithmic issues for further research and
 improvements.

\end{abstract}



\section{Introduction}\label{sec:int}
Many high-dimensional data  in real-world applications
can be modeled as data points lying close to a low-dimensional
nonlinear manifold. Discovering the structure of the manifold from
a set of data points sampled from the manifold
possibly with noise represents a very challenging unsupervised
learning problem 
\cite{dogr:02,dogr:02a,free:02,hiro:03,jema:02,koho:00,rasi:02,rosa:01,rosa:02,tero:02,tesi:00}. 
The discovered low-dimensional structures can be further 
used for classification,
clustering, outlier detection and data visualization.
Example
low-dimensional manifolds embedded in high-dimensional input spaces
include image vectors representing
the same 3D objects under different camera views and lighting conditions,
a  set of document vectors in a text corpus dealing with a specific topic,
and a set of 0-1 vectors encoding the test results on a set of multiple
choice questions for a group of students \cite{rasi:02,rosa:01,tesi:00}.
The key observation is
that the dimensions
of the embedding spaces can be very high (e.g., the number of pixels
for each images
 in the image collection,
the number of terms (words and/or phrases) in the vocabulary of the text
corpus, or the number of multiple choice questions in the test),
the intrinsic dimensionality of the data points, however, are 
rather limited due to
factors such as physical constraints and linguistic correlations.
Traditional dimension reduction techniques such as principal
component analysis and  factor 
analysis usually work 
 well when the data points lie close to a {\it linear} (affine)
subspace in the input space \cite{hatf:01}. 
They can not, in general,  discover nonlinear structures embedded in the
set of data points.

Recently, 
there have been much renewed interests in developing
efficient algorithms for constructing nonlinear low-dimensional
manifolds
from sample data points in high-dimensional spaces, emphasizing
simple algorithmic implementation and avoiding optimization
problems prone to  local minima
\cite{rosa:01,tesi:00}. Two lines of research of 
manifold learning and nonlinear
dimension reduction have emerged: one is 
exemplified by \cite{dogr:02,dogr:02a,tesi:00} 
where pairwise {\it geodesic} distances
of the data points 
with respect to the underlying manifold are estimated, and the
classical multi-dimensional scaling is used to project the data points
into a low-dimensional space that best preserves the geodesic distances.
Another line of research follows the long tradition starting with
self-organizing maps (SOM) \cite{koho:00}, principal curves/surfaces
\cite{hast:88} and topology-preserving networks \cite{mash:94}.
The key idea
is that the information about the global structure of a nonlinear
manifold can be obtained from a careful analysis of
the interactions of the {\it overlapping} local
structures. In particular, the local linear embedding (LLE) method
constructs a local geometric structure that is invariant to translations
and
orthogonal transformations in a neighborhood of each data points and
seeks
to project the data points into a low-dimensional space that best
preserves
those local geometries \cite{rosa:01,saro:02}.

Our approach draws inspiration from and
improves
upon the work in \cite{rosa:01,saro:02} which
opens up new directions in nonlinear manifold learning
with many fundamental problems requiring to be further investigated.
Our starting point is not to
consider nonlinear dimension reduction in isolation as
merely constructing a nonlinear
projection, but rather to combine it with the process of reconstruction
of the nonlinear manifold, and we argue that
the two processes interact with each
other in
a mutually reinforcing way. 
In this paper, we address two inter-related objectives
of nonlinear
structure finding: 1) to construct the so-called principal manifold
\cite{hast:88}
that goes through ``the middle'' of the data points; and 2) to find the
global coordinate system (the natural parametrization space)
that characterizes the set of data points in a
low-dimensional space. The basic idea of our approach is to use the
tangent space in the neighborhood of a data point to represent the local
geometry, and then align those local tangent spaces to construct the
global coordinate system for the nonlinear manifold.

The rest of the paper is organized as follows: in section \ref{sec:dr},
we formulate the problem of manifold learning and dimension reduction
in more precise terms, and illustrate the intricacy of the problem using
the linear case as an example. In section \ref{sec:tan}, we discuss the
issue of learning local geometry using tangent spaces, and in section
\ref{sec:align} we show how to align those local tangent spaces in order
to learn the global coordinate system of the underlying manifold.
Section \ref{sec:pm} discusses how to construct the manifold once the
global coordinate system is available. 
We call the new algorithm {\it local tangent space alignment} (LTSA)
algorithm.
In section \ref{sec:error}, we
present an error analysis of LTSA, especially illustrating
the interactions among 
curvature information embedded in the
Hessian matrices, local sampling density and noise level, and the regularity
of the Jacobi matrix. In section \ref{sec:is},
we show how the partial eigendecomposition used in global coordinate
construction can be efficiently computed. We then present a collection
of numerical experiments in section \ref{sec:num}. Section \ref{sec:con}
concludes the paper and  addresses
 several theoretical and algorithmic issues for
further research and improvements.

\section{Manifold Learning and Dimension Reduction}\label{sec:dr}
We assume that a $d$-dimensional manifold $\cal F$
embedded in an $m$-dimensional space $( d < m)$
can be represented by a function 
\[f: C \subset \R^d \rightarrow R^m,\]
 where
$C$ is a compact subset of $\R^d$
with open interior. We are given a set of data points
$x_1,\,\cdots,\,x_N$, where $x_i \in \R^m$ are sampled 
possibly with noise from the
manifold, i.e.,
\[ x_i = f(\tau_i) + \epsilon_i, \quad i=1, \dots, N,\]
where $\epsilon_i$ represents noise. By {\it dimension reduction}
we mean the estimation of the unknown lower dimensional feature vectors 
$\tau_i$'s from the $x_i$'s, i.e., the
$x_i$'s which are data points in $\R^m$ is 
(nonlinearly) projected to $\tau_i$'s
which are points in $\R^d$, with $d <m$ we realize the objective
of dimensionality reduction of the data points. By {\it manifold
learning}
we mean the reconstruction of $f$ from  the $x_i$'s, i.e., for an
arbitrary test point $\tau \in C \subset \R^d$,
we can provide an estimate of $f(\tau)$. These two problems are
inter-related,
and the solution of one leads to the solution of the other. In some
situations, dimension reduction can be the means to an end by itself,
and
it is not necessary to learn the manifold. In this paper, however, we
promote the notion that both problems are really the two sides of the
same
coin, and the best approach is not to consider each in isolation.
Before we tackle the algorithmic details,
we first want to point out that the key difficulty in manifold learning
and nonlinear dimension reduction from a sample
of data points is that the data points are {\it unorganized},
i.e., no adjacency relationship among them are known beforehand.
Otherwise,
the learning problem becomes the well-researched nonlinear
regression problem (for a more detailed
discussion, see \cite{free:02} where techniques from computational
geometry was used to solve error-free manifold learning problems).
To ease discussion, in what follows
we will call the space where the data points live
the {\it input} space, and the space into which the data points are
projected the {\it feature} space.

To illustrate
the concepts and problems we have introduced,
we consider 
the example of {\it linear}
manifold learning and {\it linear} dimension reduction.
We assume that
the set of data points are sampled from a $d$-dimensional affine
subspace,
i.e.,
\[x_i = c + U \tau_i+\epsilon_i, \quad i=1, \dots, N,\]
where $c \in \R^m, \tau_i \in \R^d$ and $\epsilon_i \in \R^m$ represents
noise.
$U \in \R^{ m\times d}$ is a matrix forms an orthonormal basis of the
affine
subspace.
Let
\[X = [x_1,\, \cdots,\,x_N],\quad T = [\tau_1,\, \cdots,\,\tau_N],\quad
E = [\epsilon_1,\, \cdots,\,\epsilon_N].\]
Then in matrix form, the data-generation model can be written as
\[X = c\,e^T+UT+E,\]
here $e$ is an $N$-dimensional column vector of all ones.
The problem of linear manifold learning 
amounts to seeking $c, U$ and $T$ to minimize the reconstruction error
$E$, i.e.,
\[ \min \|E\| = \min_{c, U, T} \;\; \|X-(c\,e^T+ UT)\|_F,\]
where $\|\cdot\|_F$
stands for the Frobenius norm of a matrix. 
This problem can be readily solved by  the
singular value decomposition (SVD)
based upon the following two observations. 

1) The norm of 
the error matrix $E$ can be reduced by removing the mean of 
the columns of $E$
from each column of $E$, 
and hence one can assume that the optimal $E$ has zero mean. 
This requirement can be fulfilled if  $c$ is chosen as the mean of $X$, 
i.e.,
$c = Xe/N \equiv \bar{x}$.

2) The low-rank matrix $UT$ is the optimal rank-$d$ approximation 
to the centered data matrix $X-\bar x e^T$.
Hence the the optimal solution is given by the 
SVD of $X-\bar x e^T$, 
\[X-\bar xe^T
  = Q\Sigma V^T, \quad P \in \R^{m \times m}, \;\;
\Sigma \in \R^{m \times N},\;\; V \in \R^{ N \times N},\]
i.e.,
$UT = Q_d\Sigma_dV_d^T,$
where $\Sigma_d=\diag(\sigma_1,\cdots,\sigma_d)$ with the $d$ largest 
singular values of $X-\bar xe^T$,
$Q_d$ and $V_d$ are the matrices of the corresponding left and right
 singular vectors, respectively.
The optimal $U^*$ is then given by $Q_d$ and the learned
linear manifold is represented by the linear function
\[ f(\tau) = \bar{x} + U^* \tau.\]
In this model, 
the coordinate matrix $T$ corresponding 
to the data matrix $X$ is given by
\[ T = (U^*)^T(X-\bar x e^T) = \diag(\sigma_1, \dots, \sigma_d) V_d^T.\]
Ideally, the dimension $d$ of the learned linear manifold 
should be chosen such that
$\sigma_{d+1} \ll \sigma_d$.

The function $f$ is not unique in the sense that it can
be reparametrized, 
i.e., the coordinate can be replaced by $\tilde \tau$ with 
a global affine transformation
$\tau = P\tilde \tau$, if we change the basis matrix $U^*$ to $U^*P$. 
What we are interested in with respect to dimension reduction is the 
low-dimensional representation of the
linear manifold in the feature space.
Therefore, without loss of generality, we can assume that 
the feature vectors are  uniformly
distributed. For a given data 
set, this  amounts to
assuming that the coordinate matrix $T$ is orthonormal in row,
i.e., $TT^T= I_d$. Hence we 
we can take $T = V_d^T$ and the linear function is now the following
\[f(\tau) = \bar{x} + U^* \diag(\sigma_1, \dots, \sigma_d)\tau.\]
For the linear case we just discussed, the problem of dimension
reduction
is solved by computing the right singular vectors $V_d$, and this can be
done without the help of the linear function $f$. Similarly, the
construction of the linear function $f$ is done by computing
$U^*$ which are just the $d$ largest left singular vectors of $X-\bar xe ^T$.

The case for nonlinear manifolds is more complicated. 
In general, 
the global nonlinear structure will have to come from local linear analysis
and alignment
\cite{rosa:01,tero:02}. 
In \cite{rosa:01}, local linear structure of the data set 
are extracted by representing each point $x_i$ as a weighted 
linear combination
of its neighbors, and 
the local weight vectors are preserved in the feature
space in order to obtain a global coordinate system.
In \cite{tero:02}, a linear alignment strategy was proposed
for aligning a general set of local linear structures.
The type of local geometric information we use is the
tangent space at a given point which is constructed from a neighborhood
of the given point. 
The local 
tangent space provides a  low-dimensional linear
approximation of the local geometric structure of  the nonlinear manifold.
What we want to preserve are the local coordinates
of the data points in the neighborhood with respect to the tangent space.
Those local tangent coordinates will be aligned in the low dimensional space
by different local affine
transformations to obtain a global coordinate system. Our alignment method
is similar in spirit to that proposed in \cite{tero:02}.
In the next section we will discuss the local tangent 
space and global alignment
which  will then be applied to data points sampled with 
noise in Section \ref{sec:align}.

\section{Local Tangent Space and Its Global Alignment}\label{sec:tan}
We assume that $\cal F$ is a $d$-dimensional manifold in a 
$m$-dimensional space 
with {\it unknown} generating function $f(\tau),\,\,\tau\in R^d$,
and we are
given a data set consists of $N$
$m$-dimensional vectors $X = [x_1, \dots, x_N], \; x_i \in \R^m$ generated from
the
following  noise-free model,
\[ x_i = f(\tau_i), \quad i=1, \dots, N, \]
where $\tau_i \in \R^d$ with $d < m$. 
The objective as we mentioned before
for nonlinear dimension reduction  is to 
reconstruct $\tau_i$'s from the corresponding  function 
values $f(\tau_i)$'s without explicitly
constructing $f$. 
Assume that the function $f$ is smooth enough, 
using first-order Taylor expansion at a fixed $\tau$, we have
\begin{eqnarray}\label{eq:taylor}
f(\bar\tau) = f(\tau) + J_f(\tau) \cdot (\bar\tau-\tau) + O(\|
\bar\tau-\tau \|^2),
\end{eqnarray}
where $J_f(\tau) \in \R^{m \times d}$ is the Jacobi matrix
of $f$ at $\tau$. If we write
the $m$ components of $f(\tau)$ as
\[ f(\tau) = \left[\begin{array}{c}
                   f_1(\tau) \\
                   \vdots \\
                   f_m(\tau)
             \end{array}\right], \quad \mbox{\rm then } \quad
 J_f(\tau) = \left[\begin{array}{ccc}
              \partial f_1/\partial \tau_1 & \cdots &
\partial f_1/\partial \tau_d \\
\vdots & \vdots &  \vdots \\
 \partial f_m/\partial \tau_1 & \cdots & \partial f_m/\partial \tau_d
\end{array}\right].\]
The tangent space ${\cal T}_\tau$ of
$f$ at $\tau$ is spanned by the $d$ column vectors
of $J_f(\tau)$ and is therefore of dimension at most $d$,
i.e., ${\cal T}_\tau =
\span (J_f(\tau))$. The
vector $\tau-\bar\tau$ gives the coordinate
of $f(\tau)$ in the affine subspace $f(\tau) + {\cal T}_\tau$.
Without knowing the function $f$,
we can not explicitly compute the Jacobi matrix $J_f(\tau)$.
However, if we know ${\cal T}_\tau$ in terms of
$Q_{\tau}$, a matrix forming an orthonormal
basis of
 ${\cal T}_\tau$, we can write
\[J_f(\tau) (\bar\tau -\tau) = Q_{\tau}\theta_\tau^*, \]
Furthermore,
\[\theta_\tau^* =  
Q_{\tau}^TJ_f(\tau) (\tau-\bar\tau)\equiv P_{\tau} (\bar\tau-\tau).
\]

The mapping from $\tau$ to $\theta_\tau^*$ represents 
a local affine transformation.
This affine transformation is unknown because we do not know the function
$f$. 
The vector $\theta_\tau^*$, however, has an
approximate $\theta_\tau$ that orthogonally projects 
$f(\bar\tau)-f(\tau)$ onto ${\cal T}_\tau$,
\begin{equation}\label{eq:P}
\theta_\tau \equiv  Q_{\tau}^T(f(\bar\tau)-f(\tau)) = \theta_\tau^* 
+ O(\|\bar\tau-\tau \|^2), 
 \end{equation}
provided $Q_\tau$ is known at each $\tau$.
Ignoring the second-order term, 
the global coordinate 
$\tau$ satisfies 
\[
\int d\tau \int_{\Omega(\tau)} \|P_\tau (\bar\tau- \tau) - \theta_\tau\|
d\bar\tau \approx 0.
\]
Here $\Omega(\tau)$ defines the neighborhood of $\tau$.
Therefore, a natural way to approximate the global coordinate is to
find a global coordinate $\tau$ and a local affine 
transformation $P_\tau$ that minimize the error function
\begin{eqnarray}\label{eq:align2}
\quad \int d\tau \int_{\Omega(\tau)} \|P_\tau (\bar\tau- \tau) - \theta_\tau\|
d\bar\tau.
\end{eqnarray}
This represents a {\it nonlinear} alignment approach for the 
dimension reduction problem (this idea will be picked up at the end of
section \ref{sec:align}).

On the other hand, a {\it linear} alignment approach can be devised as
follows. 
If $J_f(\tau)$ is of full column rank, the matrix $P_{\tau}$
should be non-singular and
\[\bar\tau-\tau \approx P_{\tau}^{-1}\theta_\tau \equiv
L_{\tau}\theta_\tau.\]
The above equation
shows that the affine transformation $L_{\tau}$ should
align this local coordinate with the {\it global} coordinate
$\tau-\bar\tau$ for $f(\tau)$. Naturally we should seek to find a global
coordinate
$\tau$ and a local affine transformation $L_{\tau}$ to
minimize
\begin{eqnarray}\label{eq:align1}
\int d\tau \int_{\Omega(\tau)} \|\bar\tau- \tau- L_{\tau}\theta_\tau\|
d\bar\tau.
\end{eqnarray}
The above amounts to matching the local geometry in
the feature space.
Notice that $\theta_\tau$ is defined by the ``known'' function 
value and the ``unknown'' 
orthogonal basis matrix $Q_\tau$ of the tangent space. 
It turns out, however,
$Q_\tau$ can be approximately determined by certain function values. 
We will discuss this approach in the next section. 
Clearly, this linear  approach is more readily 
applicable than (\ref{eq:align2}). 
Obviously, If the manifold $\cal F$ is not {\it regular}, 
i.e., the Jacobi matrix
$J_f$ is not of full column rank at some points $\tau \in C$, then 
the two minimization problems (\ref{eq:align1}) and  (\ref{eq:align2}) 
may lead to quite different solutions.

As is discussed in the linear case,
the low-dimensional feature vector
 $\tau$ is not uniquely determined
by the manifold $\cal F$.
We can reparametrize $\cal F$ using $f(g(\tau))$ where $g(\cdot)$
is a smooth $1$-to-$1$ onto mapping of $C$ to itself.
The parameterization of $\cal F$ can be fixed by requiring that
$\tau$ has a uniform distribution over $C$. This will come up as
a normalization issue in the next section.

\section{Feature Extraction through Alignment}\label{sec:align}
Now we consider how to construct the global coordinates and
local affine transformation when we are given a 
data set $X=[x_{1},\, \dots,\, x_{N}]$ sampled 
with noise from an underlying nonlinear manifold,
\[ x_i = f(\tau_i)+\epsilon_i, \quad i=1, \dots, N, \]
where $\tau_i \in \R^d$, $x_i \in \R^m$ with $d < m$. 
For each $x_i$, let
$X_i=[x_{i_1},\, \dots,\, x_{i_k}]$ be
a matrix consisting of its $k$-nearest neighbors including
$x_i$, say in terms of the
Euclidean distance.
Consider computing
the best $d$-dimensional affine subspace approximation
for the data points in  $X_i$,
\[ \min_{x, \Theta, Q} \;\; \sum_{j=1}^k \left\| x_{i_j}  -( x+ Q
\theta_j) \right\|_2^2
 =  \min_{x, \Theta, Q} \;\;  \left\| X_i-(xe^T+Q\Theta) \right\|_2^2,\]
where $Q$ is of $d$ columns and is orthonormal,
and $\Theta = [\theta_1, \dots, \theta_k]$. As is discussed in
section \ref{sec:dr},
the optimal $x$ is given by $\bar{x}_i$, the mean of all the $x_{i_j}$'s
and
the optimal $Q$ is given by $Q_i$, the $d$ left singular vectors of
$X_i(I-ee^T/k)$
corresponding to  its $d$ largest singular values,
and $\Theta$  is given by $\Theta_i$ defined as

\begin{equation}\label{eq:Theta}
\Theta_i = Q_i^TX_i(I-\frac1kee^T) = [\theta^{(i)}_1,\cdots,\theta^{(i)}_k],
\quad \theta^{(i)}_j = Q_i^T(x_{i_j}-\bar x_i).
\end{equation}
Therefore
we have
\begin{equation}\label{eq:xQ}
x_{i_j} = \bar{x}_i + Q_i \theta^{(i)}_j+\xi^{(i)}_j,
\end{equation}
where 
$\xi^{(i)}_j = (I-Q_iQ_i^T)(x_{i_j} -  \bar{x}_i)$
denotes the reconstruction error.

We now consider constructing the global coordinates $\tau_{i},
i=1, \dots, N$, in the
low-dimensional feature 
space based on the local coordinates $\theta^{(i)}_{j}$ which represents the
local geometry.
Specifically,
we want $\tau_{i_j}$ to satisfy the following set of equations, i.e.,
the global coordinates should respect the local geometry determined by
the $\theta^{(i)}_{j}$,
\begin{eqnarray}\label{eq:tau}
 \tau_{i_j} = \bar{\tau}_i + L_i \theta^{(i)}_{j}+\epsilon^{(i)}_j,
\quad j=1,\dots,k, \;\; i=1, \dots, N,
\end{eqnarray}
where $\bar{\tau}_i$ is the mean of $\tau_{i_j}, j=1, \dots, k$.
In  matrix form,
\[T_i = \frac1k T_ie e^T + L_i\Theta_i+E_i,\]
where $T_i = [ \tau_{i_1}, \dots, \tau_{i_k}]$ and
$E_i = [\epsilon^{(i)}_1,\,\cdots,\,\epsilon^{(i)}_k]$ is the local
reconstruction error matrix, and we write
\begin{eqnarray}\label{eq:Ei}
E_i = T_i(I-\frac1k e e^T)-L_i\Theta_i.
\end{eqnarray}
To preserve as much of
the {\it local} geometry in the low-dimensional 
feature space, we seek to
find $\tau_{i}$ and  the local affine transformations $L_i$ to
minimize the reconstruction errors $\epsilon^{(i)}_{j}$, i.e.,
\begin{equation}\label{eq:E}
\sum_i \| E_i \|^2\equiv \sum_i \| T_i(I-\frac1k e e^T)-L_i\Theta_i
\|^2 = \min. \end{equation}
Obviously, the optimal alignment matrix $L_i$ that minimizes the local
reconstruction error $\|E_i\|_F$ for a fixed $T_i$,
is given by
\[L_i = T_i(I-\frac{1}{k}e e^T)\Theta_i^{+} =  T_i\Theta_i^{+},\;\;
\mbox{and therefore}\;\;
E_i = T_i(I-\frac{1}{k}e e^T)(I-\Theta_i^{+}\Theta_i),\]
where $\Theta_i^{+}$ is the Moor-Penrose generalized inverse of
$\Theta_i$.
Let  $T = [ \tau_1, \dots, \tau_N]$ and
$S_i$ be the $0$-$1$  selection matrix  such that
$T S_i = T_i,$.
We then need to find $T$ to minimize the overall
reconstruction  error
\[\sum_i \|E_i\|_F^2 = \| T SW\|_F^2,\]
where $S = [S_1,\cdots,S_N]$, and $W = \diag(W_1,\cdots,W_N)$ with
\begin{equation}\label{eq:Wi}
W_i =(I-\frac{1}{k}e e^T)(I-\Theta_i^{+}\Theta_i).
\end{equation}
To uniquely determine $T$,
we will impose the constraints $TT^T = I_d$, it turns out
that the vector $e$ of all ones is an eigenvector of 
\begin{eqnarray}\label{eq:B}
B \equiv SWW^TS^T
\end{eqnarray}
corresponding to a zero eigenvalue, therefore,
the optimal $T$ is given by
the $d$ eigenvectors of the matrix $B$, corresponding
to the $2$nd to $d$+1st smallest eigenvalues of $B$.

{\sc Remark.} We now briefly discuss the {\it nonlinear} alignment
idea mentioned in (\ref{eq:align2}). In particular, in a neighborhood
of a data point $x_i$ consisting of data points 
$X_i=[x_{i_1}, \dots, x_{i_k}]$, by first order Taylor expansion, we have
\[ X_i(I-ee^T/k) \approx J_f^{(i)} T_i (I-ee^T/k).\]
Let $S_i$ be the neighborhood selection matrix as defined before, we
seek to find $J_f^{(i)} \in \R^{m \times d}$ and $T$ to minimize
\[ E(J,T) \equiv \sum_{i=1}^N \|(X-J_f^{(i)}T)S_i(I-ee^T/k)\|_F^2,\]
where $J=[ J_f^{(1)}, \dots, J_f^{(N)}]$. 
The LTSA algorithm can be considered as an approach to find an
approximate solution to the above minimization problem. We can,
however, seek to find the optimal solution of $E(J,T)$ using an
{\it alternating} least squares approach: fix $J$ minimize $E(J,T)$ 
with respect
to $T$, and fix $T$ minimize $E(J,T)$ with respect
to $J$, and so on. As an initial value to start the alternating
least squares, we can use the $T$ obtained from the LTSA algorithm. The
details of the algorithm will be presented in a separate paper.

{\sc Remark.} The minimization problem (\ref{eq:E}) needs certain constraints
(i.e., normalization conditions) to be well-posed, otherwise, one can
just choose both $T_i$ and $L_i$ to be zero. However, there are more
than one way to impose the normalization conditions. The one we
have selected, i.e., $TT^T = I_d$, is just one of the possibilities. To
illustrate the issue we look at the following minimization problem,
\[ \min_{ X,\, Y} \| X - Y A\|_F \]
The approach we have taken amounts to substituting $Y= XA^+$, and minimize
$\|X(I-A^+A)\|_F$ with the normalization condition $XX^T=I$. However,
\[ \|X - Y A\|_F = \left\|[X, Y] \left[ \begin{array}{c} I\\-A\end{array}
                                 \right] \right\|_F,\]
and we can minimize the above by imposing the normalization condition
\[[X, Y][X, Y]^T = I.\]
This nonuniqueness issue is closely related to to nonuniqueness
of the parametrization of the nonlinear manifold $f(\tau)$, which
can reparametrized as $f(\tau(\eta))$with a 1-to-1 
mapping $\tau(\eta)$.


\section{Constructing Principal Manifolds}\label{sec:pm}
Once the global coordinates $\tau_i$ are computed for each of the
data points $x_i$, we can apply some non-parametric
regression methods such as local polynomial regression
to  $\{(\tau_i, x_i)\}_{i=1}^N$ to construct the principal manifold
underlying the set of points $x_i$.
Here each of the component functions $f_j(\tau)$ can be constructed
separately, 
for example, we have used the simple {\tt loess} function \cite{veri:99}
in some of
our experiments for generating the principal manifolds.

In general, when the low-dimensional
 coordinates $\tau_i$ are available, 
we can construct an mapping 
from the $\tau$-space (feature space) to the $x$-space 
(input space) as follows.

1. For each fixed $\tau$, let $\tau_i$ be the nearest neighbor
(i.e., $\|\tau-\tau_i\| \leq \|\tau-\tau_j\|,$ for $j \neq i$). Define
\[\theta = L_i^{-1}(\tau-\bar \tau_i),\]
where $\bar \tau_i$ be the mean of the feature vectors in a
neighbor to which $\tau_i$ belong.

2. Back in the input space, we define
\[x = \bar x_i+Q_i\theta.\]

Let us define by $g:\,\,\tau\rightarrow x$ the resulted mapping,
\begin{eqnarray}\label{eq:g}
g(\tau) = \bar x_i+Q_iL_i^{-1}(\tau-\bar \tau_i). 
\end{eqnarray}
To distinguish the computed coordinates $\tau_i$ from the 
exact ones, in the rest of this paper,
we denote by $\tau^*_i$ the exact coordinate, i.e.,
\begin{eqnarray}\label{eq:xf}
x_i = f(\tau^*_i)+\epsilon^*_i.
\end{eqnarray}
Obviously,  
the errors of the reconstructed manifold represented by $g$
depend on the sample errors $\epsilon^*_i$, 
the local tangent subspace reconstruction errors $\xi^{(i)}_j$, 
and the alignment errors $\epsilon^{(i)}_j$.
The following result show that this dependence is linear.
\begin{theorem}\label{thm:error}
Let  
$\epsilon^*_i = x_i-f(\tau^*_i)$, 
$\xi^{(i)}_j=(I-Q_iQ_i^T)(x_i-\bar x_i)$, and 
$\epsilon_i = \tau_i-\bar \tau_i-L_iQ_i^T(x_i-\bar x_i)$. Then
\[\|g(\tau_i)-f(\tau_i^*)\|_2\leq 
\|\epsilon^*_i\|_2+\|\xi_i\|_2+ \|L_i^{-1}\epsilon_i \|_2.\]
\end{theorem}
\begin{proof}
Substituting $L_i^{-1}(\tau_{i}-\bar\tau_i) = L_i^{-1}\epsilon_i+Q_i^T(x_i-\bar x_i)$ into (\ref{eq:g}) gives
\begin{eqnarray}
g(\tau_i) 
&=& \bar x_i+Q_iL_i^{-1}(\tau_{i}-\bar\tau_i)\nonumber\\
&=& \bar x_i+Q_iQ_i^T(x_i-\bar x_i)+Q_iL_i^{-1}\epsilon^{(i)}_1.\nonumber 
\end{eqnarray}
Because $Q_iQ_i^T(x_i-\bar x_i)=x_i-\bar x_i-\xi^{(i)}_j $, 
we obtain that
\begin{eqnarray}
g(\tau_i) 
&=&x_i-\xi^{(i)}_j+Q_iL_i^{-1}\epsilon^{(i)}_1 \nonumber\\
&=& f(\tau^*_i) +\epsilon^*_i -\xi^{(i)}_j+Q_iL_i^{-1}\epsilon^{(i)}_1.\nonumber
\end{eqnarray}
Therefore we have
\[\|g(\tau_i)-f(\tau_i^*)\|_2\leq 
\|\epsilon^*_i\|_2+\|\xi_i\|_2+ \|L_i^{-1}\epsilon_i \|_2,\]
completing the proof.
\end{proof}

In the next section, we will give a detailed error analysis 
to estimate the errors of alignment and tangent space approximation
in terms of the noise, the geometric properties of the generating
function $f$ and the density of the generating coordinates $\tau^*_i$.

\section{Error Analysis}\label{sec:error}
As is mentioned in the previous section, 
we assume that that the data points are generated by
\[ x_i = f(\tau^*_i) + \epsilon^*_i, \quad i=1, \dots, N.\]
For each $x_i$, let
$X_i=[x_{i_1},\, \dots,\, x_{i_k}]$ be
a matrix consisting of its $k$-nearest neighbors including
$x_i$ in terms of the
Euclidean distance.
Similar to $E_i$  defined in (\ref{eq:Ei}),
we denote by $E^*_i$ the corresponding local noise matrix,
$E^*_i = [ \epsilon^*_{i_1}, \dots, \epsilon^*_{i_k}]$. 
The low-dimensional embedding coordinate matrix computed
by the LTSA algorithm is denoted by $T=[\tau_1,
\dots, \tau_N]$.
We first
present a result that bounds $\|E_i\|$ in terms of $\|E^*_i\|.$

\begin{theorem}\label{thm:E}
Assume $T^*=[\tau_1^*,
\dots, \tau_N^*]$ satisfies $(T^*)^TT^*=U_d$.
Let  $\bar\tau_i$ be the mean of $\tau_{i_1},\,\cdots,\,\tau_{i_k}$,
Denote  $P_i = Q_i^TJ_f(\bar\tau^*_i)$ and
$H_{f_\ell}(\bar\tau^*_i)$ the Hessian matrix of
the $\ell$-th component function of $f$.
If        
the $P_i$'s are nonsingular, then
\[\|E_i\|_F\leq \|P_i^{-1}\|_F(\delta_i+\|E_i^*\|_F),\]
where $\delta_i$ is defined by
\[\delta_i^2 = \sum_{\ell=1}^m
\sum_{j=1}^k\|H_{f_\ell}(\bar\tau^*_i)\|_2^2\,
  \|\tau^*_{i_j}-\bar\tau^*_i\|_2^4\]
Furthermore, if each neighborhood is of size $O(\eta)$, then
$\|E\| \leq \|P_i^{-1}\|_F\|E^*\| +O(\eta^2)$.
\end{theorem}
\begin{proof}
First by  (\ref{eq:Ei}), we have
\begin{eqnarray}\label{eq:Ei1}
 E_i = T_i(I-\frac1k e e^T)-L_i\Theta_i = (T_i-L_iQ_i^TX_i)(I-\frac1k e
e^T).
\end{eqnarray}
To represent $X_i$ in terms of the Jacobi matrix of $f$,
we assume that $f$ is smooth enough and use Taylor expansion at
$\bar \tau^*_i$, the mean of the $k$ neighbors of $\tau^*_i$, we have
\[ x_{i_j} = f(\bar \tau^*_i)+J_i(\tau^*_{i_j}-\bar\tau^*_i)
+ \delta^{(i)}_j+ \epsilon_{i_j},\]
where
$J_i = J_f(\bar\tau^*_i)$ and
$\delta^{(i)}_j$ represents the remainder term beyond the first
order expansion, in particular,
its $\ell$-th components can be approximately written as (using
second order approximation),
\[ \delta^{(i)}_{\ell,j}\approx \frac{1}{2}
(\tau^*_{i_j}-\bar\tau^*_i)^TH_{f_\ell}(\bar\tau^*_i)(\tau^*_{i_j}-\bar\tau^*_i)\]
with the Hessian matrix $H_{f_\ell}(\bar\tau^*_i)$ of the $\ell$-th
component
function $f_\ell$ of $f$ at $\bar\tau^*_i$.
We have in matrix form,
\[X_i =  f(\bar \tau^*_i)e^T + J_iT^*_i(I-\frac1k e e^T) + \Delta_i +
E^*_i\]
with $\Delta_i = [\delta^{(i)}_1,\cdots,\delta^{(i)}_k]$.
Multiplying by the centering matrix $I-\frac1k e e^T$ gives
\begin{eqnarray}\label{eq:Xi}
X_i(I-\frac1k e e^T) = (J_iT^*_i+\Delta_i+E^*_i)(I-\frac1k e e^T).
\end{eqnarray}
Substituting (\ref{eq:Xi}) into (\ref{eq:Ei1}) and denoting $P_i =
Q_i^TJ_i$,  we obtain that
\begin{eqnarray}\label{eq:Ei2}
\quad\quad E_i = (T_i-L_iP_iT^*_i-L_i Q_i^T (\Delta_i+E^*_i))(I-\frac1k
e e^T).
\end{eqnarray}
For any $\tilde T$ satisfying the orthogonal condition $\tilde T\tilde
T^T=I_d$ and any $\tilde L_i$,
we also have the similar expression of (\ref{eq:Ei2}) for $\tilde T_i$
and $\tilde L_i$.
Note that $T$ and $L_i$, $i=1,\cdots,N$, minimize the overall
reconstruction error,
$\|E\|_F\leq \|\tilde E\|_F$.
Setting $\tilde T=T^*$ and $\tilde L_i = P_i^{-1}$, we obtain the upper
bound
\[\|E_i\|_F\leq \|P_i^{-1}\|_2(\|\Delta_i\|_F+\|E^*_i\|_F).\]
We estimate the norm $\|\Delta_i\|_F$ by ignoring the higher order
terms, and
obtain that
\[\|\Delta_i\|_F^2\leq
\sum_{\ell=1}^m
\sum_{j=1}^k\|H_{f_\ell}(\bar\tau^*_i)\|_2^2\,\|\tau^*_{i_j}-\bar\tau^*_i\|_2^4
=\delta^2,\]
completing the proof.
\end{proof}

The non-singularity of the
matrix  $P_i$ requires that  the 
Jacobi matrix $J_i$ be of full column
rank and the two subspaces
$\span(J_i)$ and the $d$ largest left singular vector space $\span(Q_i)$ 
are not orthogonal to each other. We now give a quantitative measurement
of the non-singularity of $P_i$.
\begin{theorem}\label{thm:P}
Let $\sigma_d(\tilde J_i)$ be the $d$-th singular value of 
$\tilde J_i\equiv J_iT^*_i(I-\frac1k e e^T)$, and denote
$\alpha_i = 4(\|E^*_i\|_F+\delta_i)/\sigma_d(\tilde J_i)$ 
with $\delta_i$ defined in 
Theorem \ref{thm:E}.
Then
\[\| P_i^{-1}\|_F\leq (1+\alpha_i^2)^{1/2}\|J_i^+\|_F.\]
\end{theorem}
\begin{proof}
The proof is simple. Let
$\tilde J_i=U_J\Sigma_JV_J^T$
 be the SVD of the matrix 
$\tilde J_i$. 
By (\ref{eq:Xi}) and perturbation bounds for
singular subspaces \cite[Theorem 8.6.5]{govl:96},
the singular vector matrix $Q_i$ can be expressed as
\begin{eqnarray}\label{eq:Qi}
Q_i = (U_J+U_J^\bot H)(I+H_i^TH_i)^{-1/2}
\end{eqnarray}
with 
\[\|H_i\|_F\leq \frac{4}{\sigma_d(\tilde J_i)}
  \Big( \|E^*\|_F+  \|\Delta_i\|_F) \Big)\leq\alpha_i,\]
where $\sigma_d(\tilde J_i)$ is the $d$-largest singular value of 
$\tilde J_i$.
On the other hand, from the SVD of $\tilde J_i$, we have 
$J_iT^*_iV_J =  U_J\Sigma_J$, which gives 
\[J_i = U_J\Sigma_J\big(T^*_iV_J\big)^{-1}.\]
It follows that
\[ P_i = Q_i^TJ_i = (I+H_i^TH_i)^{-1/2}\Sigma_J \big(T^*_iV_J\big)^{-1} =
(I+H_i^TH_i)^{-1/2}U_J^TJ_i. \]
Therefore we have
\[\| P_i^{-1}\|_F\leq (1+\|H_i\|_F^2)^{1/2}\|J_i^+\|_F,\]
completing the proof.
\end{proof}

The degree of non-singularity of $J_i$ is determined by the curvature 
of the manifold and 
the rotation of the singular subspace is mainly affected by the sample 
noises $\epsilon_j$'s and the neighborhood structure of $x_i$'s. The
above error bounds clearly show that reconstruction accuracy will
suffer if the manifold underlying the data set has singular or near-singular
points. This phenomenon will be illustrated in the numerical examples
in section \ref{sec:num}.
Finally, we give an error upper bound for the tangent subspace approximation.
\begin{theorem}\label{thm:TS}
Let $\cond(\tilde J_i) = \sigma_1(\tilde J_i)/\sigma_d(\tilde J_i)$ 
be the  spectrum condition number of the $d$-column matrix $\tilde J_i$.
Then
\[\|X_i-(\bar x_i e^T+Q_i\Theta_i)\|_F
\leq  \left(1+4(1+\alpha_i^2)\cond(\tilde J_i)\right)(\|E^*_i\|_F+\delta_i).\]
\end{theorem}
\begin{proof}
By (\ref{eq:Xi}), we write 
\[(I-Q_iQ_i^T)X(I-\frac{1}{k}ee^T) = (I-Q_iQ_i^T)\tilde J_i+\tilde \Delta_i,\]
with $\|\tilde \Delta_i\|_F\leq \|E^*_i\|_F+\delta_i$.
To estimate $\|(I-Q_iQ_i^T)\tilde J_i\|_F$, we use the expression
(\ref{eq:Qi}) to obtain 
\begin{eqnarray*}
(I-Q_iQ_i^T)\tilde J_i 
&=& U
\left(\left(\begin{array}{cc}I\\O\end{array}\right)
     -\left(\begin{array}{cc}I\\H_i\end{array}\right)(I+H_i^TH_i)^{-1}
\right)\Sigma_JV_J^T\\
&=& U\left(\begin{array}{cc}H_i^T\\-I\end{array}\right)
H_i(I+H_i^TH_i)^{-1}\Sigma_JV_J^T.
\end{eqnarray*}
Taking norms gives that
\[\|(I-Q_iQ_i^T)\tilde J_i\|_F\leq
(1+\|H_i\|_2^2)\|H_i\|_F\|\tilde J_i\|_2
\leq 4(1+\alpha_i^2)(\|E^*_i\|_F+\delta_i)\cond(\tilde J_i).
\]
The result required follows.
\end{proof}
The above results show that the accurate determination of the local tangent
space is dependent on several factors: curvature information embedded in the
Hessian matrices, local sampling density and noise level, and the regularity
of the Jacobi matrix.

\section{Numerical Computation Issues}\label{sec:is}
One major computational cost of LTSA involves the computation of the
smallest eigenvectors of the  
symmetric positive semi-defined matrix $B$ defined in (\ref{eq:B}).
$B$ in general will be quite sparse because of the local nature
of the construction of the neighborhoods. Algorithms for computing
a subset of the eigenvectors for large and/or sparse matrices are
based
on computing 
projections of $B$
onto a sequence of  Krylov subspaces of the form
\[ K_p(B, v_0) = \span \{ v_0, Bv_0, B^2v_0, \dots, B^{p-1}v_0\}, \]
for some initial vectors $v_0$ \cite{govl:96}.
Hence the computation of matrix-vector
multiplications $B x$ needs to be done efficiently. 
Because of the special nature of $B$, $Bx$ can be computed neighborhood
by neighborhood without explicitly forming $B$,
\[ B x  = S_1W_1W_1^T S_1^T x + \cdots + S_N W_NW_N^T S_N^T x,\]
where as defined in (\ref{eq:Wi}),
\[W_i = (I-\frac{1}{k})(I-\Theta_i^+\Theta_i).\]
Each term in the above summation only involves the
$x_i$'s in one neighborhood.

The matrix $\Theta_i^+\Theta_i$ in the right factor of $W_i$ is the orthogonal projector onto the 
subspace spanned by the rows of $(\Theta_i)$. 
If the SVD of $X_i-\bar x_ie^T = Q^{(i)}\Sigma^{(i)}(H^{(i)})^T$ is available, 
the orthogonal projector is given by
$\Theta_i^+\Theta_i = H_iH_i^T$, where $H_i$ is the submatrix of first $d$ columns of $H^{(i)}$.
Otherwise, one can compute the QR decomposition $\Theta_i^T=H_iR_i$ 
of $\Theta_i^T$ 
and obtain $\Theta_i^+\Theta_i = H_iH_i^T$ \cite{govl:96}. 
Clearly, we have $H^Te = 0$ because $\Theta_ie=0$.
Then we can rewrite $W_i$ as 
\[W_i = I-\frac{1}{k}ee^T-H_iH_i^T = I-[e/\sqrt{k},\, H_i][e/\sqrt{k},\, H_i]^T \equiv I-G_iG_i^T.\] 
It is a orthogonal projector onto the null space spanned by the rows of $\Theta_i$ and $e^T/\sqrt{k}$.
Therefore the matrix-vector product $ y=S_iW_iW_i^T S_i^T x$ can be easily computed as follows: 
denote by $I_i=\{i_1,\cdots,i_k\}$ the set of indices
for the $k$ nearest-neighbors of $x_i$, then the $j$-th element of $y$ is zero for $j\notin I_i$ 
and
\[y(I_i) = x(I_i)-G_i(G_i^Tx(I_i)). \]
Here $y(I_i)=[y_{i_1},\cdots,y_{i_k}]^T$ denotes the section of $y$ 
determined by the neighborhood set $I_i$.

If one needs to compute the $d$ 
smallest eigenvectors that are orthogonal to $e$ by applying
some eigen-solver with an explicitly 
formed  $B$. 
The matrix $B$ can be computed by carrying out a
partial local  summation as follows
\begin{eqnarray}\label{eq:B1}
B(I_i,I_i)\,\, \leftarrow\,\,B(I_i,I_i) + I-G_iG_i^T,\quad i=1,\cdots,N
\end{eqnarray}
with initial $B=0$.

Now we are ready to present
our Local Tangent Space Alignment (LTSA) algorithm.

\bigskip

\begin{center}
\fbox{\parbox{11cm}{ {\bf Algorithm LTSA} (Local Tangent Space Alignment). 
Given $N$ $m$-dimensional points 
sampled possibly 
with noise from an underlying $d$-dimensional manifold, 
this algorithm produces $N$ $d$-dimensional coordinates 
$T \in \R^{d\times N}$ for the manifold
constructed from $k$ local nearest neighbors.

\begin{itemize}
\item[{\bf Step 1.}] [Extracting local information.] For each $i=1,\cdots,N$,
  \begin{itemize}
  \item[{\bf 1.1}] Determine $k$ nearest neighbors $x_{i_j}$ of $x_i, \, j=1, \dots, k$.
  \item[{\bf 1.2}] Compute the $d$ largest eigenvectors $g_1,\cdots,g_d$ of the correlation matrix
                   $(X_i-\bar x_i e^T)^T(X_i-\bar x_i e^T)$, and set 
         \[G_i = [e/\sqrt{k},\, g_1,\,\cdots,\,g_d].\]
  \end{itemize}

\item[{\bf Step 2.}] [Constructing alingment matrix.] Form the matrix $B$ by locally summing
       (\ref{eq:B1}) if a direct eigen-solver will be used. Otherwise
implement a routine that computes matrix-vector multiplication $Bu$ for
an arbitrary vector $u$.
      
\item[{\bf Step 3.}] [Aligning global cordinates.]
Compute 
the $d+1$ smallest eigenvectors of $B$ and pick up the eigenvector
matrix $[u_2,\,\cdots,\,u_{d+1}]$ corresponding 
to the 2nd to $d+1$st smallest eigenvalues,
and set $T = [u_2,\,\cdots,\,u_{d+1}]^T$.
\end{itemize}
}}
\end{center}

\bigskip

\section{Experimental Results}\label{sec:num}
In this section, we present several numerical examples to illustrate the
performance of our LTSA algorithm. The test data sets include curves in
2D/3D Euclidean spaces, and surfaces in 3D Euclidean spaces. Especially,
we take a closer look at the effects of singular points of a manifold
and the interaction of noise levels and sample density. To show
that our algorithm can also handle data points in high-dimensional spaces,
we also consider curves and surfaces in Euclidean spaces with dimension
equal to $100$ and an image
data set with dimension $4096$. 

\begin{figure}[t]
\centerline{
    \mbox{\psfig{file=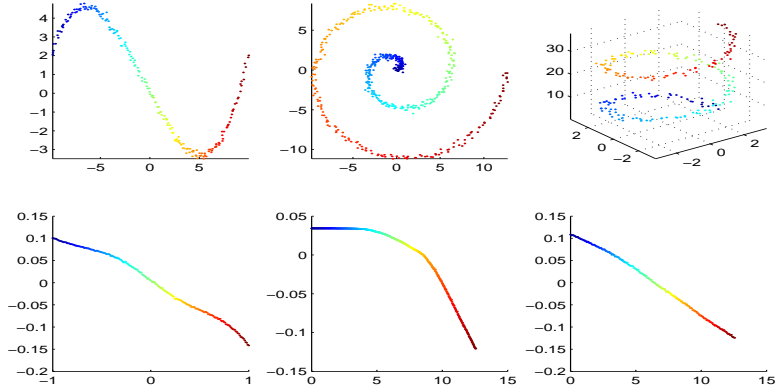,width=4.4in,height=2.2in}}
}
\caption{sample data points with noise
from various 1-D manifolds (top) and coordinates 
of computed $\tau_i$ via exact $\tau_i^*$  (bottom).}
\label{fig:1D}
\end{figure}

First we test our LTSA method for 1D manifolds (curves) in both 2D and 3D. 
For a given 1D manifold $f(\tau)$ with uniformly sampled coordinates
$\tau_1^*,\cdots,\tau_N^*$ in a fixed interval, 
we add Gaussian noise to obtain the data set $\{x_i\}$ as follows,
\[x_i = f(\tau_i^*)+\eta \, {\tt randn}(m,1),\]
where $m=2,3$ is the dimension of the input space, and
{\tt randn} is Matlab's standard normal distribution.
In Figure \ref{fig:1D}, in the first row from left to right,
we plot the color-coded sample data points corresponding to the following
three one-variable functions
\[\begin{array}{rcllr}
f(\tau) &=& (10\tau,\,\,10\tau^3+2\tau^2-10\tau)^T, & \tau\in[-1,\,\, 1],& \eta=0.1,\\
f(\tau) &=& (\tau\cos(\tau),\,\,\tau\sin(\tau))^T,  & \tau\in[0,\,\, 4\pi],& \eta=0.2,\\
f(\tau) &=& (3\cos(\tau),\,\,3\sin(\tau),\,\,3\tau)^T,& \tau\in[0,\,\, 4\pi],& \eta=0.2.
\end{array}
\]
In the second row, we
plot $\tau_i^*$ against $\tau_i$, where $\tau_i$'s are the computed
coordinates by LTSA. Ideally, the $(\tau_i^*,  \tau_i)$ should form
a straight line with either a $\pi/4$ or $-\pi/4$ slope.

\begin{figure}[t]
\centerline{
    \mbox{\psfig{file=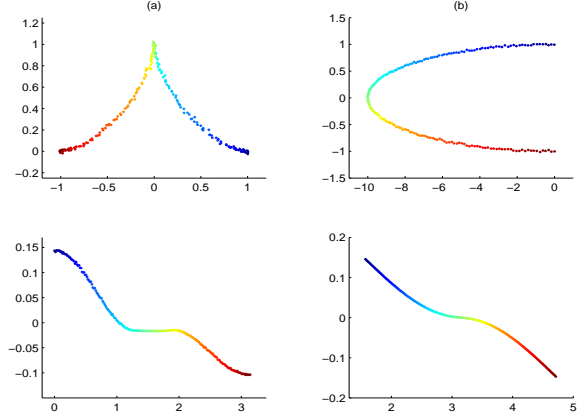,width=3.0in,height=2.2in}}
}
\caption{1-D manifolds with singular points (left) and 
corresponding coordinates $\tau_i$ via exact $\tau_i^*$ (right).}
\label{fig:singular}
\end{figure}

It is also important
to better understand the failure modes of LTSA, and ultimately 
to identify  conditions under which LTSA can truly uncover
the hidden nonlinear structures of the data points.
As we have shown in the error analysis
in section \ref{sec:error}, it will be difficult to 
align the local tangent information $\Theta_i$ 
if some of the $P_i$'s defined in (\ref{eq:P}) are
close to be singular. One effect of this
is that the computed 
coordinates $\tau_i$ and its neighbors may be 
compressed together. To clearly demonstrate  this
phenomenon, we consider the following function,
\[ f(\tau) = [\cos^3(\tau),\,\,\sin^3(\tau)]^T, \quad \tau\in[0,\, \pi].\]
The Jacobi matrix (now a single vector since $d=1$) given by
\[J_f(\tau) = 1.5\sin(2\tau) [-\cos(\tau),\,\, \sin(\tau)]^T \]
is equal to zero at $\tau=\pi/2$. In that case 
the $\theta$-vector  $\Theta_i$ defined in (\ref{eq:Theta})
will be computed poorly in the presence of noise. 
Usually the corresponding $\Theta_i$ will be small which also results in 
small $\tau_i$ and the neighbors
of $\tau_i$ will also be small. 
In the first column 
of Fig \ref{fig:singular}, we 
plot the computed results for this 1-D curve. 
We see clearly near the singular point $\tau= \pi/2$ 
the computed $\tau_i$'s become very
small, all compressed to a small interval around zero.
In the second column
of Fig \ref{fig:singular}, we examine another
1D curve defined by
\[f(\tau) = [10\cos(\tau),\,\, \sin(\tau)]^T,\quad \tau\in[\pi/2,\,\,3\pi/2].\]
We notice that similar phenomenon also occurs 
near the point $\tau= \pi$ where
the {\it curvature} of the curve is large, the computed $\tau_i$'s near
the corresponding point
also become very
small, clustering around zero. 

\begin{figure}[t]
\centerline{
    \mbox{\psfig{file=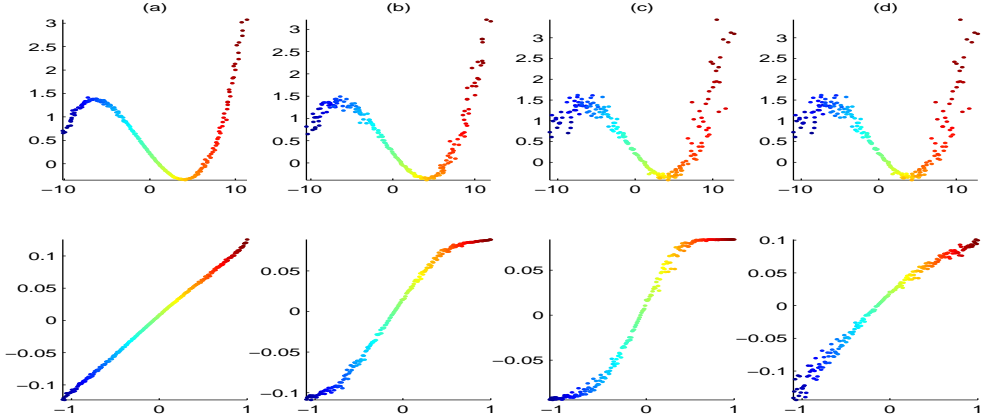,width=5.2in,height=2.2in}}
}
\caption{1-D manifolds with different noise levels (top) 
and computed coordinates $\tau_i$ vs. exact $\tau_i^*$  (bottom).}
\label{fig:1Dnoise}
\end{figure}

Next we look at the issues of 
the interaction of sampling density and noise levels. 
If there are large noises around $f(\tau_i)$
relative to the sampling density near $f(\tau_i)$, the resulting 
centered local data matrix $X_i(I-\frac{1}{k}ee^T)$ 
will not be able to provide a good local tangent space, i.e.,  
$X_i(I-\frac{1}{k}ee^T)$ will have singular values 
$\sigma_d$ and $\sigma_{d+1}$ that are close to each other. 
This will result in a nearly singular matrix $P_i=Q_i^TJ_i$,
and when plotting $\tau_i^*$ against $\tau_i$, we will see
the phenomenon of the computed coordinates $\tau_i$ getting
compressed, similar to the case 
when the generating function $f(\tau)$ has  singular and/or near-singular
points.
However, in this case, the result 
 can  usually be improved by increasing the number of 
neighbors used for producing the shifted 
matrix $X_i(I-\frac{1}{k}ee^T)$. In Fig \ref{fig:1Dnoise}, 
we plot the computed results for the generating function 
\[f(\tau) = 3\tau^3+2\tau^2-2\tau,\quad \tau\in[-1.1,\,1].\]
The data set is generated by adding noise in a relative fashion,
\[x_i = f(\tau_i)(1 + \eta\epsilon_i)\]
with normally distributed $\epsilon_i$. 
The first three columns in Fig \ref{fig:1Dnoise} 
correspond to the noise levels
$\eta=0.01$, $\eta=0.03$, and $\eta=0.05$, respectively. 
For the three data sets, 
We use the same number of neighbors,
$k=10$. With the increasing noise level $\eta$, 
the computed $\tau_i$'s get expressed at points 
with relatively large noise. The quality of the computed $\tau_i$'s can be
improved if we increase the number of neighbors as is shown
on the column (d) in Fig \ref{fig:1Dnoise}. 
The improved result is for the same data set in column (c) with
$k=20$  used.

As we have shown in Fig \ref{fig:1Dnoise} (column (d)), 
different neighborhood size $k$ 
will produce different embedding results. 
In general, $k$ should be chosen to match the sampling density, noise
level and the curvature at each
data points  so as to 
extract an accurate local tangent space. Too few neighbors used may result 
in a
rank-deficient tangent space leading to 
over-fitting, while too large a  neighborhood 
will introduce too much bias and the computed  tangent space
will not match the local geometry well.
It is therefore worthy of considering variable number of neighbors 
that are adaptively chosen at 
each data point.
Fortunately, our LTSA algorithm seems 
to be less sensitive to the choice of $k$ than LLE does as will be shown
later. 

\begin{figure}[t]
\centerline{
    \mbox{\psfig{file=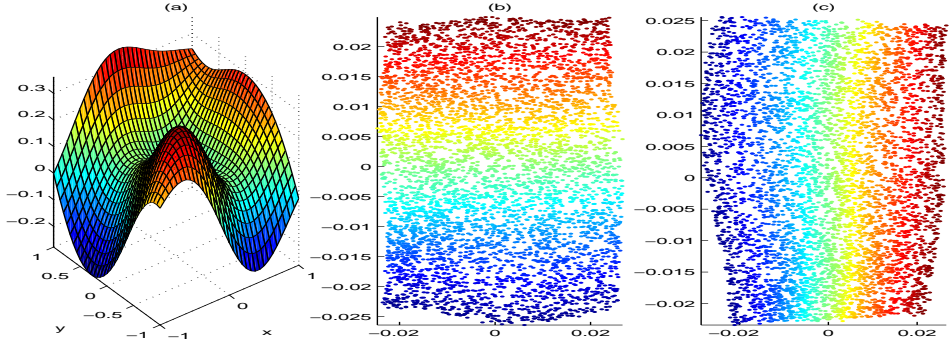,width=5.0in,height=1.8in}}}
\caption{2-D manifold in a 100-D space generated  by 3-D peak function: (a) 3-D peak curve, 
(b) coordinates for orthogonal transformed  manifold, (c) coordinates for affine transformed  manifold. }
\label{fig:100D}
\end{figure}

We now apply LTSA to a $2$-D manifold embedded in a 100 dimensional space.
The data points are  generated as follows.
First we generate $N=5000$ $3$D points, 
\[x_i = (t_i,\,s_i,\,h(t_i,s_i))^T + 0.01\eta_i\] 
with $t_i$ and $s_i$ uniformly distributed in the interval $[-1,\,\,1]$,
the $\eta_i$'s are standard normal.
The $h(t,s)$ is a peak function defined by
\[h(t,s) = 0.3(1-t)^2e^{-t^2-(s+1)^2} 
             - (0.2t-t^3-s^5)e^{-t^2-s^2}
             - 0.1e^{-(t+1)^2 - s^2}.\]
This function is plotted in the left of Figure \ref{fig:100D}. 
We generate two kinds of data points $x_i^Q$ and $x_i^H$ in 100D space,
\[ x_i^Q = Qx_i,\quad x_i^H = Hx_i,\]
where $Q$ is a random orthogonal matrix resulting in
an orthogonal transformation
and $H$ a matrix with 
its
singular values  uniformly distributed in $(0,\,1)$ resulting in an
affine transformation. 
Figure \ref{fig:100D} plots the coordinates for $x_i^Q$ (middle) 
and $x_i^H$ (right).

\begin{figure}[t] 
\centerline{
\mbox{\psfig{file=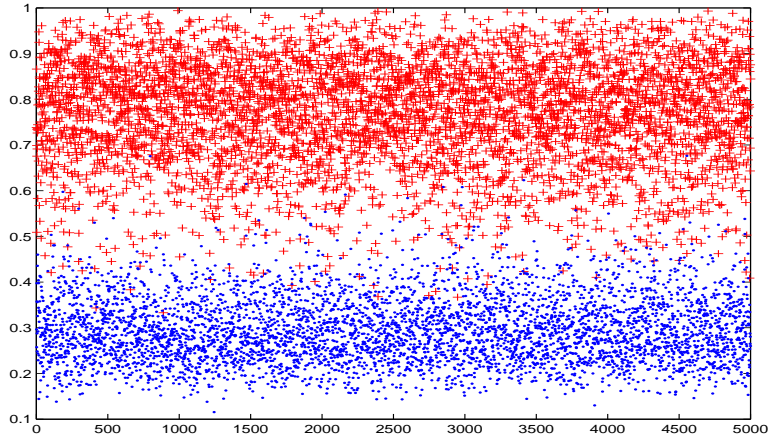,height=2.3in,width=4.in}}}
\caption{Singular value ratios $\rho_i^{(1)}$ ($+$-dots) and
$\rho_i^{(2)}$ ($\cdot$-dots).}
\label{fig:Eratio}
\end{figure}

One advantage of LTSA over LLE is that
using LTSA we can potentially
detect the intrinsic dimension of the underlying manifold by analyzing
the local tangent space structure. In particular, we can examine the 
distribution of the singular
values of the local data matrix $X_i$ consisting of the data points in the
neighborhood of each data point $x_i$. If the manifold is of dimension
$d$, then  $X_i$ will be close to a rank-$d$ matrix. We illustrate
this point below.
The data points are $x^Q_i$ of the $2$D peak manifold in the $100$D
space.
For each local data matrix $X_i$,
let $\sigma_{j,i}$ be the $j$-the singular value of the centered matrix
$X_i(I-\frac{1}{k}ee^T)$.
Define the ratios
\[\rho_i^{(j)} = \frac{\sigma_{j+1,i}}{\sigma_{j,i}}.\]
In Fig \ref{fig:Eratio}, we plot the rations $\rho_i^{(1)}$ and  
$\rho_i^{(2)}$.
It clearly shows the feature space should be $2$-dimensional.
  
\begin{figure}[t]
\centerline{
\mbox{\psfig{file=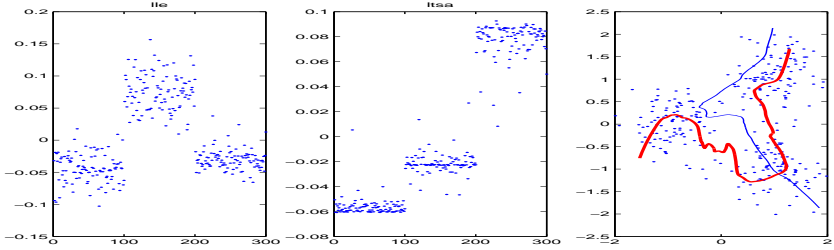,height=1.3in,width=4.4in}}}
\caption{(left) Global coordinates by LLE, (middle)
global coordinates by LTSA, (right) Three
Gaussian data with principal curves}
\label{fi:3gaussian}
\end{figure}

Next, we discuss the issue of how to use the global
coordinates $\tau_i$'s as a means for clustering the data points $x_i$'s.
The situation is illustrated by Figure \ref{fi:3gaussian}. The data
set consists of three bivariate Gaussians with covariance
matrices $0.2I_2$ and mean vectors located at $[1, 1], [1, -1], [-1, 0]$. 
There are $100$ sample points from each Gaussian. The
thick curve on the right panel represents the principal curve computed
by LTSA and the thin curve by LLE. It is seen
that  the thick curve goes through
each of the Gaussians in turn, and the corresponding global coordinates
(plotted in the middle panel) clearly separate the three Gaussians. LLE
did not perform as well, mixing two of the Gaussians.

\begin{figure}[t]
\centerline{
\mbox{\psfig{file=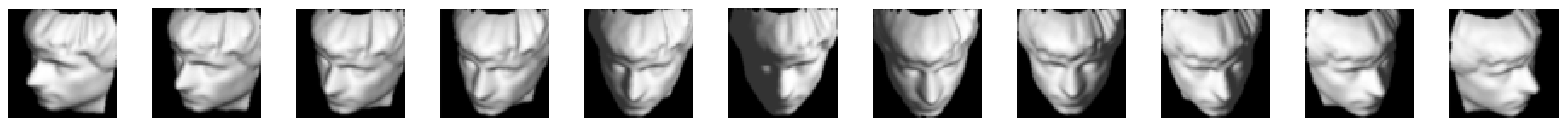,height=0.36in,width=5.0in}}
}
\centerline{
\mbox{\psfig{file=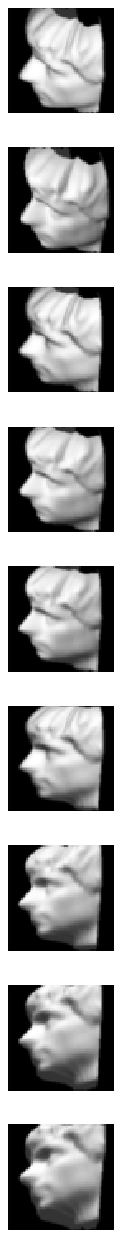,height=3.4in,width=0.4in}
      \psfig{file=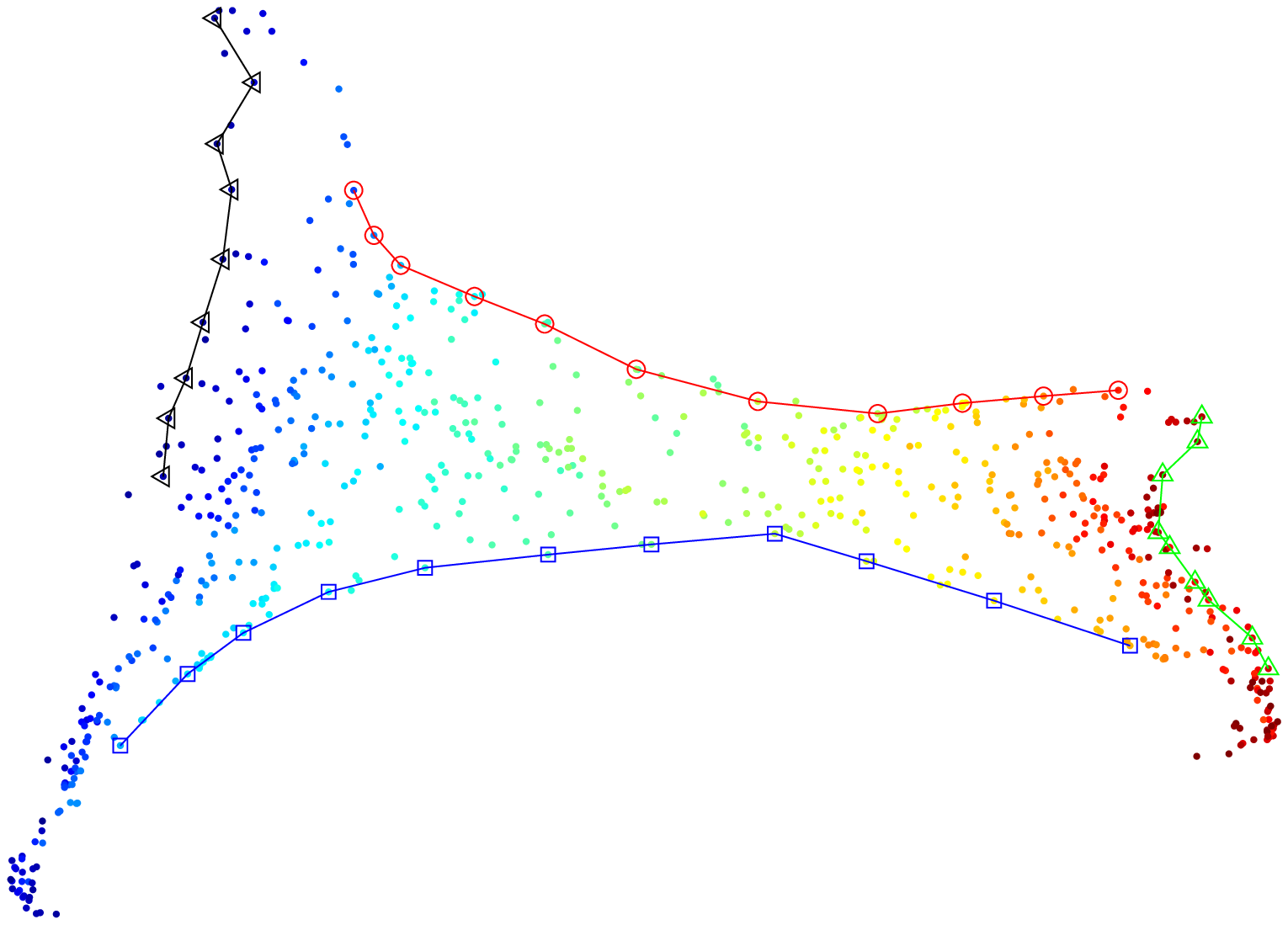,height=3.4in,width=4.1in}
      \psfig{file=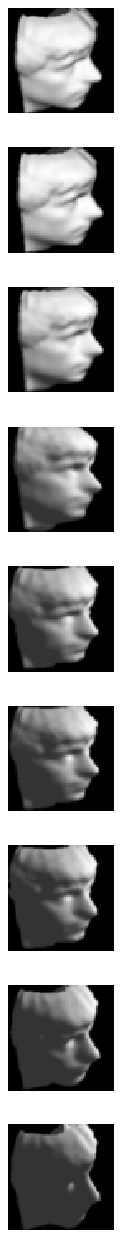,height=3.4in,width=0.4in}
}}
\centerline{\mbox{\psfig{file=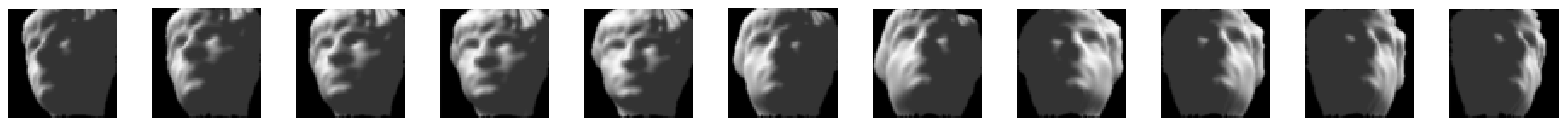,height=0.36in,width=5.0in}
}}
\caption{Coordinates computed by Algorithm LTSA with $k=12$ neighbors (middle)
and images corresponding to the points on the bound lines (top, bottom,left, and right) 
Left.}
\label{fig:face}
\end{figure}

Last we look at
the results of applying LTSA algorithm 
to the face image data set \cite{tesi:00}. The data set
consists of a
sequence $698$ $64$-by-$64$ pixel images of a
face rendered under various pose and lighting conditions. Each image is
converted to an $m=4096$ dimensional image vector. We apply LTSA with
$k=12$ neighbors and $d=2$. The constructed coordinates are plotted in the
middle of Figure \ref{fig:face}. We also extracted four paths along the
boundaries of of the set of the 2D coordinates, and display the corresponding
images along each path. It can be seen that the computed 2D coordinates
do capture very well the pose and lighting variations in a continuous way.

\section{Conclusions and Feature Works}\label{sec:con}
In this paper, we proposed a new algorithm (LTSA) for
nonlinear manifold learning and nonlinear dimension reduction.
The key techniques we used are construction of local tangent spaces to
represent local geometry and the global alignment of the local
tangent spaces to obtain the global coordinate system for the
underlying manifold. We provide some careful error analysis to 
exhibit the interplay of approximation accuracy, sampling density,
noise level and curvature structure of the manifold. In the following,
we list several issues that deserve further investigation.

1. To make LTSA (similarly LLE)  more robust
against noise, we need to resolve the issue
where several of the smallest eigenvalues of
$B$ are about the same magnitude. This
can be clearly seen when the manifold itself consists of several 
disjoint components. If this is the case, one needs to break the
matrix $B$ into several diagonal blocks, and apply LTSA to each
of the block. However, with noise, the situation becomes far more
complicated, several eigenvectors corresponding to near-zero eigenvalues
can mix together, the information of the global coordinates seems to
be contained in the eigen-subspace, but how to unscramble the eigenvectors
to extract the global coordinate information needs more careful analysis
of the eigenvector matrix of $B$ and various models of the noise.
Some preliminary results on this problem 
have been presented in \cite{pope:01}.

2. The selection of the set of points to estimate the local tangent space
is very crucial to the success of the algorithm. Ideally, we want this
set of points to be close to the tangent space. However, with noise and/or
at the points where the curvature of the manifold is large, this is not
an easy task. One line of ideas is to do some preprocessing of the data
points to construct some {\it restricted} local neighborhoods. For
example, one can first compute the minimum Euclidean spanning tree for
the data set, and restrict the neighbors of each point to those that
are linked by the branches of the spanning tree. This idea has been
applied in self-organizing map \cite{koho:00}. Another idea is 
to use iterative-refinement, combining the computed  $\tau_i$'s with
the $x_i$'s for neighborhood construction in another round
of nonlinear projection. The rationale is that
$\tau_i$'s as the computed global coordinates of the nonlinear manifold
may give a better measure of the local geometry.

3. A {\it discrete} version of the manifold learning can be formulated by
considering the data points as the vertices of an undirected graph
 \cite{mash:94}. A
quantization of the global coordinates specifies the adjacency relation
of those vertices, and manifold learning becomes  discovering
whether an edge should
be created between a pair of vertices or not so that the resulting
vertex neighbors resemble those of the manifold. We need to investigate
a proper formulation of the problem and the related optimization methods.

4. From a statistical point of view, it is also of great interest to
investigate more precise formulation of the error
model and the associated 
consistency issues and convergence rate as the sample size 
goes to infinity. The learn-ability
of the nonlinear manifold also
depends on the
 sampling density of the data points. 
Some of the results in non-parametric regression and statistical
learning theory will
be helpful to pursue this line of research.

\end{document}